# Statistical Augmentation of a Chinese Machine-Readable Dictionary

**Pascale Fung**
*Columbia University*
Computer Science Department
New York, NY 10027
USA
pascale@cs.columbia.edu

**Dekai Wu**
*HKUST*
Department of Computer Science
University of Science & Technology
Clear Water Bay, Hong Kong
dekai@cs.ust.hk

## Abstract

We describe a method of using statistically-collected Chinese character groups from a corpus to augment a Chinese dictionary. The method is particularly useful for extracting domain-specific and regional words not readily available in machine-readable dictionaries. Output was evaluated both using human evaluators and against a previously available dictionary. We also evaluated performance improvement in automatic Chinese tokenization. Results show that our method outputs legitimate words, acronymic constructions, idioms, names and titles, as well as technical compounds, many of which were lacking from the original dictionary.

## 1   Introduction

Finding new lexical entries for Chinese is hampered by a particularly obscure distinction between characters, morphemes, words, and compounds. Even in Indo-European text where words can be separated by spaces, no absolute criteria are known for deciding whether a collocation constitutes a compound word. Chinese defies such distinctions yet more strongly. Characters in Chinese (and in some other Asian languages such as Japanese and Korean) are not separated by spaces to delimit words; nor do characters give morphological hints as to word boundaries. Each single character carries a meaning and can be ambiguous; most are many-way polysemous or homonymous.

Some characteristics of Chinese words are nonetheless clear. A word in Chinese is usually a bigram (two character word), a unigram, a trigram, or a 4-gram. Function words are often unigrams, and $n$-grams with $n > 4$ usually are specific idioms. According to the *Frequency Dictionary of Modern Chinese* (FDMC 1986), among the top 9000 most frequent words, 26.7% are unigrams, 69.8% are bigrams, 2.7% are trigrams, 0.007% 4-grams, and 0.0002% 5-grams. Another study (Liu 1987) showed that in general, 75% of Chinese words are bigrams, 14% trigrams, 6% $n$-grams with $n > 3$.

Inadequate dictionaries have become the major bottleneck to Chinese natural language processing. Broad coverage is even more essential than with Indo-European languages, because not even the most basic lexicosyntactic analysis can proceed without first identifying the word boundaries. Thus a significant number of models for *tokenizing* or *segmenting* Chinese have recently been proposed, using either rule-based or statistical methods (Chiang *et al.* 1992; Lin *et al.* 1992; Chang & Chen 1993; Lin *et al.* 1993; Wu & Tseng 1993; Sproat *et al.* 1994). But all of these approaches rely primarily upon dictionary lookup of the potential segments; in spite of experimental heuristics for handling unknown words in the input text, accuracy is seriously degraded when dictionary entries are missing.

Tokenization problems are aggravated by text in specialized domains. Such documents typically contain a high percentage of technical or regional terms that are not found in the tokenizer's dictionary (machine-readable Chinese dictionaries for specialized domains are not readily available). Most effective tokenizers have domain-specific words added manually to the dictionary. Such manual strategies are too tedious and inefficient in general.

This paper discusses a fully automatic statistical tool that extracts words from an untokenized Chinese text, creating new dictionary entries. In addition, it is desirable to identify regional and domain-specific technical terms that are likely to appear repeatedly in a large corpus. We extended and re-targeted a tool originally designed for extracting English compounds and collocations, Xtract, to find words in Chinese. We call the resulting tool **CXtract**. Words found by CXtract are used to augment our dictionary.

In the following sections, we first describe the modifications in CXtract for finding Chinese words, and the corpus used for training. The resulting words and collocations are evaluated by human evaluators, and recall and precision are measured against the tokens in the training set. The significance of evaluated results will be discussed. Finally, we discuss a preliminary evaluation of the improvement in tokenization performance arising from the output of our tool.

## 2   A Collocation Extraction Tool

Xtract was originally developed by Smadja (1993) to extract collocations in an English text. It consists of a package of software tools used to find likely co-occurring word groups by statistical analysis.

In the first stage of Xtract, all frequent bigrams are found. These bigram words are permitted to occur within a window of 10 positions, specifically, at distance between -5 to 5 relative to each other. A threshold is set on the frequency, to discard unreliable bigrams. The remaining bigrams constitute part of the output from Xtract (along with the output from the second stage).

The second stage looks at a tagged corpus and to find collocations of involving more than two words—up to ten—using the bigram words found in the first stage as anchors. Again, a frequency threshold is set to discard unreliable collocations.

Xtract's output consists of two types of collocations. In the simpler case, a collocation is an adjacent word sequence such as "stock market" (extracted from the Wall Street Journal). More general collocations permit flexible distances between two word groups, as in "make a ... decision".

For our purpose, we were interested in looking for adjacent character groups without distances between the groups. We postulated that, just as "stock market" could be regarded as a compound word, we would discover that frequently appearing continuous character groups are likely to be words in Chinese. We were also interested in looking for multi-word collocations in Chinese since these would presumably give us many technical and regional terms.

Because Xtract was originally developed for English, many capabilities for handling non-alphabetic languages were lacking. We extended Xtract to process character-based Chinese texts without tags. Various stages of the software were also modified to deal with untagged texts.

Other parametric modifications arose from the difference between the distribution of characters that make up Chinese words, versus the words that make up English compounds. For example, the frequency threshold for finding reliable bigrams is different because CXtract returns far more Chinese character bigrams than English word bigrams returned by Xtract.

# 3  Experiment I: Dictionary Augmentation

Our experiments were aimed at determining whether our statistically-generated output contains legitimate words. We are using text from (the Chinese part of) the HKUST English-Chinese Parallel Bilingual Corpus (Wu 1994), specifically, transcriptions of the parliamentary proceedings of the Legislative Council. The transcribed Chinese is formalized literary Cantonese that is closer to Mandarin than conversational Cantonese. However, more vocabulary is preserved from classical literary Chinese than in Mandarin, which affects the ratio of bigrams to other words.

Evaluation of legitimate Chinese words is not trivial. It is straightforward to evaluate those outputs that can be found in a machine-readable dictionary such as the one used by the tokenizer. However, for unknown words, the only evaluation criterion is human judgement. We evaluated the output lexical items from CXtract by both methods.

## 3.1  Procedure

For Experiment I, we used a portion of the corpus containing about 585,000 untokenized Chinese characters (which turned out to hold about 400 thousand Chinese words after tokenization). The experiment was carried out as follows:

1: A dictionary of all unigrams of characters found in the text was composed. One example is the character 立 (*li*) which can mean "stand" or "establish" by itself.

2: From the unigram list, we found all the bigrams associated with each unigram and obtained a list of all bigrams found.

3: We kept only bigrams which occur significantly more than chance expectation, and which appear in a rigid way (Smadja 1993). This yields a list of possible bigrams and most frequent relative distance between the two characters. The distances are kept between -5 and 5 as in Xtract since this ultimately gives collocations of lengths up to 10, which we found sufficient for Chinese.

4: From this bigram list, we extracted only those bigrams in which the two characters occur adjacently. We assumed such bigrams to be Chinese words. For 立 (*li*), one output bigram was 立法 (*li fa*) which means "legislative", a legitimate word.

5: Using all bigrams (adjacent and non-adjacent) from (3), we extracted words and collocations of lengths greater than two. Outputs with frequency less than 8 were discarded.

6: We divided the output from (5) into lists of trigrams, 4-grams, 5-grams, 6-grams, and $m$-grams where $m > 6$. One of the trigrams, for example, is 立法局 (*li fa ju*) which means "Legislative Council" and is another legitimate word.

## 3.2 Results

A portion of the list of bigrams obtained from (4) is shown in Figure 1. We obtained 1695 such bigrams after thresholding.

Part of the output from (5) is shown in Figure 2. The first and the last numbers on each line is the frequency for the occurrence of the $n$-grams.

Parts of the output from (6) are shown in Figures 3, 4, and 5.

| 立法 | legislative | 人士 | personage | 政府 | government |
| 提供 | supply | 發展 | develop(ment) | 意見 | opinion |
| 部份 | partial | 包括 | include | 教育 | educate(-tion) |
| 要求 | demand | 政治 | politics | 而且 | moreover |
| 制度 | system | 研究 | research | 一般 | in general |
| 市民 | citizen | 組別 | group/category | 一個 | one *classifier* |
| 法律 | law | 條例 | regulation | 報告 | report |
| 缺乏 | lack/deprivation | 關係 | relation | 如果 | if/suppose |
| 修訂 | revise(-sion) | 表示 | express/indicate | 關注 | attention |
| 任何 | any | 反映 | reaction | 一九 | nineteen |
| 利益 | benefit | 部門 | department | 增加 | increase |
| 公平 | fair | 本身 | itself | 精神 | spirit |
| 工作 | job | 特別 | special | 數字 | number |

Figure 1: Part of the bigram output, with glosses

## 3.3 Human evaluation

For the first part of the precision evaluation, we relied on human native speakers of Mandarin and Cantonese. Many of the output words, especially domain-specific words and collocations, were not found in the tokenizer dictionary. Most importantly, we are interested in the percentage of output sequences that are legitimate words that can be used to augment the tokenizer.

Four evaluators were instructed to mark whether each entry of the bigram and trigram outputs was a word. The criterion they used was that a word must be able to stand by itself and does not need context to have a meaning. To judge whether 4-gram, 5-gram, 6-gram and $m$-gram outputs were words, the evaluators were told to consider an entry a word if it was a sequence of shorter words that taken together held a conventional meaning, and did not require any additional characters to complete its meaning.

| Freq | Collocation |
|---|---|
| 277 | .................... 副主席先生, ................ |
| 337 | .................... 政府 .................... |
| 50 | .................... 利益 .................... |
| 30 | ................ 政府可否告知本局 ................ |
| 16 | .................... 第五號報告書 ................ |
| 27 | .................... 明白 .................... |
| 20 | ....... 私事 ........ 政府當局 .. 訂究證文 . , . 私 .......... |
| 16 | .............. 謹此陳辭,支持動議。 ........ 議員致辭 ..... |
| 12 | .................... 工商界 .................... |

Figure 2: Part of the CXtract output

| | | |
|---|---|---|
| 不公平 | 不知交和 | 供某人政 a21 |
| 中小學 | 中英雙方 | 兩個市政局 |
| 公務員 | 中國政府 | 券及期貨事 |
| 及他方 | 公共援助 | 委員會提出 |
| 及其他 | 分區直選 | 的生活方式 |
| 尤其是 | 支持動議 | 的投票制度 |
| 大多數 | 文件所載 | 的選舉制度 |
| 大故是 | 人權法案 | 的選舉制度 |
| 大家都 | 夾心階層 | 建議修訂內 |
| 大部份 | 和會計師 | 很大的缺失 |
| 工市政 | 的代表性 | 施政報告中 |
| 立法局 | 的是人一 | 施政報告內 |
| 任何人 | 直選議席 | 副主席先生 |
| 全日制 | 社會人士 | 商務委員會 |
| 工商界 | 社會福利 | 專責委員會 |

Figure 3: Part of the trigram, 4-gram and 5-gram output

Besides *correct*, the evaluators were given three other categories to place the *n*-grams. *Wrong* means the entry had no meaning or an incomplete meaning. *Unsure* means the evaluator was unsure. Note that the percentage in this category is not insignificant, indicating the difficulty of defining Chinese word boundaries even by native speakers. *Punctuation* means one or more of the characters was punctuation or ASCII markup.

Tables 1 and 2 show the results of the human evaluations. The *Precision* column gives

| | |
|---|---|
| 一九九七年後 | after the year 1997 |
| 人權法案條例 | Human Rights Bill |
| 中英聯合聲明 | Sino-British Joint Declaration |
| 以及他們所提 | and as they mentioned |
| 加上九七年能 | additionally in 1997 we can |
| 本人謹此陳辭 | I hereby move that |
| 刑事罪行條例 | Criminal Law Bill |
| 至一九九七年 | until the year 1997 |
| 我們必須審刻 | we must examine |
| 我們應該予支 | we should allocate |
| 我們應該糾支 | we should correct |
| 見是民上是內 | *error* |
| 兩個市政局的 | two Urban Council's |
| 券及期貨事務 | *error* and Commodity Affairs |
| 動議投贊成票 | move to cast a supporting vote |
| 第五號報告書 | The Number 5 Report |
| 港特別行政區 | Hong Kong Special Administrative Region |
| 機場核心工程 | Airport Core Project |
| 選舉委員會的 | of the Elective Committee |
| 總督施政報告 | The Executive Report of the Governor |
| 謹此提出動議 | hereby beg to move |

Figure 4: Part of the 6-gram output, with glosses

the percentage correct over total $n$-grams in that category.

We found some discrepancies between evaluators on the evaluation of *correct* and *unsure* categories. Most of these cases arose when an $n$-gram included the possessive 的 (*de*), or the copula 是 (*shi*). We also found some disagreement between evaluators from mainland China and those from Hong Kong, particular in recognizing literary idioms.

The average precision of the bigram output was 78.13%. The average trigram precision was 31.3%; 4-gram precision 36.75%; 5-gram precision 49.7%; 6-gram precision 55.2%; and the average $m$-gram precision was 54.09%.

### 3.4 Dictionary/text evaluation

The second part of the evaluation was to compare our output words with the words actually present in the text. This gives the recall and precision of our output with respect to the training corpus. Unfortunately, the training corpus is untokenized and too large to tokenize by hand. We therefore estimated the words in the training corpus by passing it through an automatic tokenizer based on the BDC dictionary (BDC 1992). Note that this dictionary's

```
一 九 九 二 年 十 月
一 九 九 二 年 六 月
一 九 九 二 至 九 三 年 度
一 九 九 五 年 選 舉
已 是 中 英 聯 合 聲 明
支 持 麥 理 覺 議 員 的
支 持 麥 理 覺 議 員 的 修 訂
支 持 麥 理 覺 議 員 的 修 訂 動 議
支 持 麥 理 覺 議 員 的 修 訂 動 議
支 持 麥 理 覺 議 員 的 動 議
由 現 在 至 一 九 九 七 年
在 委 員 會 審 議 階 段
多 及 立 法 局 改 革 委 員 會
委 員 會 審 議 階 段
持 麥 理 覺 議 員 的
政 府 可 否 告 知 本 局
修 訂 動 議 經 向 委 員 會 提 出
選 舉 事 宜 專 責 委 員 會 報 告 書
選 舉 委 員 會 的 成 員
選 舉 委 員 會 的 組 成
總 督 在 施 政 報 告
總 督 的 施 政 報 告
總 督 施 政 報 告 中
總 督 商 務 委 員 會
總 督 彭 定 康 先 生 支
議 員 的 修 訂 動 議
議 員 致 辭 的 譯 文
```

Figure 5: Part of the $m$-gram output

entries were not derived from material related to our corpus. The tokens in the original tokenized text were again sorted into unique bigrams, trigrams, 4-grams, 5-grams, 6-grams, and $m$-grams with $m > 6$. Table 3 summarizes the precision, recall, and augmentation of our output compared to the words in the text as determined by the automatic tokenizer. *Precision* is the percentage of sequences found by CXtract that were actually words in the text. *Recall* is the percentage of words in the text that were actually found by CXtract. *Augmentation* is the percentage of new words found by CXtract that were judged to be correct by human evaluators but were not in the dictionary.

The recall is low because CXtract does not include $n$-grams with frequency lower than 8. However, we obtained 467 legitimate words or collocations to be added to the dictionary

Table 1: Human Evaluation of the Bigram Output Precision

| Evaluator | wrong | unsure | punctuation | precision |
|---|---|---|---|---|
| A | 339 | 53 | 111 | |
| | 20% | 3.1% | 6.5% | 75.2% |
| B | 264 | 31 | 111 | |
| | 15.6% | 1.8% | 6.5% | 81.4% |
| C | 269 | 118 | 111 | |
| | 15.87% | 6.96% | 6.5% | 75.6% |
| D | 289 | 23 | 111 | |
| | 17% | 1.4% | 6.5% | 80.3% |

Table 2: Human Evaluation of $n$-gram Output Precision

| Evaluator | $n$ | wrong | correct | unsure | punctuation | precision |
|---|---|---|---|---|---|---|
| A | 3 | 205 | 81 | 33 | 25 | 23.5% |
| | 4 | 98 | 89 | 5 | 20 | 44.1% |
| | 5 | 33 | 48 | 1 | 6 | 54.5% |
| | 6 | 9 | 32 | 5 | 1 | 68% |
| | $m$ | 14 | 32 | 3 | 0 | 65.3% |
| D | 3 | 296 | 101 | 23 | 25 | 29.4% |
| | 4 | 102 | 75 | 2 | 23 | 37.13% |
| | 5 | 36 | 44 | 2 | 6 | 50% |
| | 6 | 20 | 26 | 0 | 1 | 55.3% |
| | $m$ | 18 | 27 | 0 | 4 | 55.1% |
| E | 3 | 168 | 134 | 16 | 26 | 39% |
| | 4 | 89 | 81 | 10 | 22 | 40.1% |
| | 5 | 29 | 44 | 5 | 10 | 50% |
| | 6 | 10 | 0 | 11 | 1 | 53.2% |
| | $m$ | 12 | 0 | 7 | 4 | 53.06% |
| C | 3 | 210 | 112 | 0 | 22 | 32.6% |
| | 4 | 131 | 52 | 0 | 19 | 25.7% |
| | 5 | 40 | 39 | 0 | 9 | 44.3% |
| | 6 | 25 | 21 | 0 | 1 | 44.7% |
| | $m$ | 24 | 21 | 0 | 4 | 42.9% |

and the total augmentation is 5.73%. The overall precision is 59.3%.

However, we believe the frequency threshold of 8 was too low and the 585K character size of the corpus was too small. Most of the "garbage" output had low frequencies. The precision rate can be improved by using a larger data base and raising the threshold as in

Table 3: Precision, Recall and Augmentation of CXtract Output

| $n$ | Token types | CXtract | Precision | Recall | Augmentation |
|---|---|---|---|---|---|
| 2 | 6475 | 1201 | 852 (70.9%) | 662 (10.2%) | 190 (2.9%) |
| 3 | 721 | 344 | 115 (33.4%) | 10 (1.4%) | 105 (14.6%) |
| 4 | 911 | 202 | 75 (37.1%) | 7 (0.008%) | 68 (7.5%) |
| 5 | 38 | 88 | 43 (48.9%) | 0 (0%) | 43 (113.2%) |
| 6 | 7 | 47 | 29 (61.7%) | 0 (0%) | 29 (414.2%) |
| $m$ | 4 | 49 | 32 (65.3%) | 0 (0%) | 32 (800%) |
| Total | 8156 | 1931 | 1146 (59.3%) | 769 (14%) | 467 (5.73%) |

Experiment II.

In the following sections, we discuss the significance of the evaluated results.

## 3.5 Bigrams are mostly words

Using human evaluation, we found that 78% of the bigrams extracted by our tool were legitimate words (as compared with $70.9\% + 2.9\% = 73.8\%$ by evaluation against the automatic tokenizer's output). Of all $n$-gram classes, the evaluators were least unsure of correctness for bigrams, although quite a few classical Chinese terms were difficult for some of the evaluators.

Since the corpus is an official transcript of formal debates, we find many terms from classical Chinese which are not in the machine-readable dictionary, such as 謹此 (*jin ci*, "I hereby").

Some of the bigrams are acronymic abbreviations of longer terms that are also domain specific and not generally found in a dictionary. For example, 中英 (*zhong ying*) is derived from 中國,英國 (*zhong guo, ying guo*), meaning Sino-British. This acronymic derivation process is highly productive in Chinese.

## 3.6 The whole is greater than the sum of parts

What is a *legitimate* word in Chinese? To the average Chinese reader, it has to do with the vocabulary and usage patterns s/he acquired. It is sometimes disputable whether 立法局 (*li fa ju*, "Legislative Council") constitutes one word or two. But for the purposes of a machine translation system, for example, the word 局 (*ju*) may be individually translated not only into "Council" but also "Station", as in 警察局 (*jing cha ju*, "Police Station"). So we might incorrectly get "Legislative Station". On the other hand, 立法局 (*li fa ju*) as a

single lexical item always maps to "Legislative Council"

Another example is 大部份 (*da bu fen*) which means "the majority". Our dictionary omits this and the resulting tokenization is 大 (*da*, "big") and 部份 (*bu fen*, "part/partial"). It is clear that "majority" is a better translation than "big part".

### 3.7 Domain specific compounds

Many of the $n$-grams for $n > 3$ found by CXtract are domain-specific compounds. For example, due to the topics of discussion in the proceedings, "the year 1997" appears very frequently.

Longer terms are frequently abbreviated into words of three or more characters. For example, 中英雙方 (*zhong ying shuang fang*) means "bilateral Sino-British", and 中英聯合聲明 (*zhong ying lian he sheng ming*) means "Sino-British Joint Declaration". Various titles, committee names, council names, projects, treaties, and joint-declarations are also found by our tool. Examples are shown in Figure 6.

Although many of the technical terms are a collocation of different words and sometimes acceptable word boundaries are found by the tokenizer, it is preferable that these terms be treated as single lexical items for purposes of machine translation, information retrieval, or spoken language processing.

### 3.8 Idioms and *cheng yu*

From $n$-gram output where $n > 3$, we find many idiomatic constructions that could be tokenized into series of shorter words. In Chinese especially, there are many four character words which form a special idiomatic class known as 成語 (*cheng yu*). There are dictionaries of *cheng yu* with all or nearly all entries being four character idioms (e.g., Chen & Chen 1983). In the training corpus we used, we discovered new *cheng yu* that were invented to describe a new concept. For example, 夾心階層 (*jia xin jie ceng*) means "sandwich class" and is a metaphorical term for families who are not well off but with income just barely too high to qualify for welfare assistance. Such invented terms are highly domain dependent, as are the usage frequencies of established *cheng yu*.

### 3.9 Names

Tokenizing Chinese names is a difficult task (Sproat *et al.* 1994) because Chinese names start with a unigram or bigram family name, and are followed by a given name freely composed of one or two characters. The given name usually holds some meaning, making it hard to distinguish names from other words. For names, we do not want to tokenize them into separate characters. In a large corpus, names are often frequently repeated. For example, in our data, the names of some parliamentary members are extracted by our tool as separate lexical items. Examples are shown in Figure 7. The last two characters of each example are the person's title.

| | |
|---|---|
| 白 皮 書 | White Paper |
| 行 政 局 | Executive Council |
| 工 商 界 | Industry and Trade |
| 立 法 局 | Legislative Council |
| 保 安 司 | Security Secretary |
| 財 政 司 | Financial Secretary |
| 經 濟 司 | Economics Secretary |
| 聯 合 聲 明 | Joint Declaration |
| 一 九 九 七 年 | the year 1997 |
| 立 法 局 選 舉 | Election of the Urban Council |
| 商 務 委 員 會 | Commerce and Trading Committees |
| 專 責 委 員 會 | Select Committees |
| 選 舉 委 員 會 | Elective Committees |
| 醫 院 管 理 局 | Hospital Administration Committee |
| 警 務 處 處 長 | Police Chief |
| 人 權 法 案 條 例 | Human Rights Bill |
| 中 英 聯 合 聲 明 | Sino-British Joint Declaration |
| 刑 事 罪 行 條 例 | Criminal Law Bill |
| 多 議 席 單 票 制 | many-seats one-vote system |
| 港 特 別 行 政 區 | Hong Kong Special Administrative Region |
| 機 場 核 心 工 程 | Airport Core Project |
| 計 由 現 在 至 一 九 九 七 年 | counting from now until the year 1997 |
| 由 現 在 至 一 九 九 七 年 | from now until the year 1997 |
| 委 員 會 審 議 階 段 | examining period of the committee |
| 教 育 統 籌 司 員 會 | Education Commission |
| 總 督 商 務 委 員 會 | Trading Committee of the Governor |

Figure 6: Some domain specific terms found by CXtract, with glosses

## 4  Experiment II: Tokenization Improvement

Given the significant percentage of augmented words in Experiment I, we can see that many entries could be added to the dictionary used for automatic tokenization.

In the next stage of our work, we used a larger portion of the corpus to obtain more

Chinese words and collocations, and with higher reliability. These items were converted into dictionary format along with their frequency information.

To obtain a baseline performance, the tokenizer was tested with the original dictionary on two separate test sets. It was then tested with the statistically-augmented dictionary on the same test sets. Each of the tokenization outputs was evaluated by three human evaluators.

```
李 柱 銘 議 員
李 華 明 議 員
林 鉅 津 議 員
彭 定 康 先 生
馮 檢 基 議 員
黃 宏 發 議 員
詹 培 忠 議 員
劉 慧 卿 議 員
譚 耀 宗 議 員
涂 謹 申 議 員
周 梁 淑 怡 議 員
麥 理 覺 議 員 的
```

Figure 7: Some names and titles found by CXtract

## 4.1 Procedure

As training data we used about 2 million Chinese characters taken from the same HKUST corpus. This is about 4 times the size used in Experiment I. The tokenizer we used employs a maximal matching strategy with frequency preferences.

The original dictionary for the tokenizer holds 104,501 entries and lacks many of the domain-specific and regional words found in the corpus.

From the first stage of CXtract, we obtained 4,196 unique adjacent bigrams. From the second stage, we filtered out any CXtract output that occurred less than 11 times and obtained 7,121 lexical candidates. Additional filtering constraints on high-frequency characters were also imposed on all candidates.[1] After all automatic filtering, we were left with 5,554 new dictionary entries.

Since the original dictionary entries employed frequency categories of integer value from 1 to 5, we converted the frequency for each lexical item from the second stage output to this same range by scaling. The adjacent bigrams from the first stage were assigned the frequency number 1 (the lowest priority).

---

[1] A refined version of the linguistic filtering is discussed in Wu & Fung (1994).

The converted CXtract outputs with frequency information were appended to the dictionary. Some of the appended items were already in the dictionary. In this case, the tokenization process uses the higher frequency between the original dictionary entry and the the CXtract-generated entry.

The total number of entries in the augmented dictionary is 110,055, an increase of 5.3% over the original dictionary size of 104,501.

### 4.2 Results

Two independent test sets of sentences were drawn from the corpus by random sampling with replacement. TESTSET I contained 300 sentences, and TESTSET II contained 200 sentences. Both sets contain unretouched sentences with occasional noise and a large proportion of unknown words, i.e., words not present in the original dictionary. (Sentences in the corpus are heuristically determined.)

Each test set was tokenized twice. *Baseline* is the tokenization produced using the original dictionary only. *Augmented* is the tokenization produced using the dictionary augmented by CXtract.

Three human evaluators evaluated each of the test sets on both baseline and augmented tokenizations. Two types of errors were counted: false joins and false breaks. A false join occurs where there should have been a boundary between the characters, and a false break occurs where the characters should have been linked. A conservative evaluation method was used, where the evaluators were told to not to mark errors when they felt that multiple tokenization alternatives were acceptable.

The results are shown in Tables 4, 5, and 6. Baseline error is computed as the ratio of the number of errors in the baseline tokenization to the total number of tokens found. Augmented error is the ratio of the total number of errors in the augmented tokenization to the total number of tokens found.

Our baseline rates demonstrate how sensitive tokenization performance is to dictionary coverage. The accuracy rate of 76% is extremely low compared with other reported percentages which generally fall around the 90's (Chiang *et al.* 1992; Lin *et al.* 1992; Chang & Chen 1993; Lin *et al.* 1993). We believe that this reflects the tailoring of dictionaries to the particular domains and genres on which tokenization accuracies are reported. Our experiment, on the other hand, reflects a more realistic situation where the dictionary and text are derived from completely independent sources, leading to a very high proportion of missing words. Under these realistic conditions, CXtract has shown enormous utility. The error reduction rate of 33% was far beyond our initial expectations.

Table 4: Result of TESTSET I - 300 sentences

| Eval-uator | # tokens | Baseline # errors | Error rate | Accuracy | # tokens | Augmented # errors | Error rate | Accuracy |
|---|---|---|---|---|---|---|---|---|
| A | 4194 | 1128 | 27% | 73% | 3893 | 731 | 19% | 81% |
| F | 4194 | 1145 | 27% | 73% | 3893 | 713 | 18% | 82% |
| G | 4194 | 1202 | 29% | 71% | 3893 | 702 | 18% | 82% |

Table 5: Result of TESTSET II - 200 sentences

| Eval-uator | # tokens | Baseline # errors | Error rate | Accuracy | # tokens | Augmented # errors | Error rate | Accuracy |
|---|---|---|---|---|---|---|---|---|
| A | 3083 | 737 | 24% | 76% | 2890 | 375 | 13% | 87% |
| H | 3083 | 489 | 16% | 84% | 2890 | 322 | 11% | 89% |
| I | 3083 | 545 | 18% | 82% | 2890 | 339 | 12% | 88% |

Table 6: Average accuracy and error rate over all evaluators and test sets

| Experiment | Total # tokens | Average error | Error rate | Accuracy |
|---|---|---|---|---|
| Baseline | 7277 | 1749 | 24% | 76% |
| Augmented | 6783 | 1061 | 16% | 84% |

## 5 Conclusion

We have presented a statistical tool, CXtract, that identifies words without supervision on untagged Chinese text. Many domain-specific and regional words, names, titles, compounds, and idioms that were not found in our machine-readable dictionary were automatically extracted by our tool. These lexical items were used to augment the dictionary and to improve tokenization.

The output was evaluated both by human evaluators and by comparison against dictionary entries. We have also shown that the output of our tool helped improve a Chinese tokenizer performance from 76% to 84%, with an error reduction rate of 33%.

# 6  Acknowledgement

We would like to thank Kathleen McKeown for her support and advice, and Frank Smadja and Chilin Shih for helpful pointers. We would also like to thank our evaluators, Philip Chan, Eva Fong, Duanyang Guo, Zhe Li, Cindy Ng, Derek Ngok, Xuanyin Xia, and Michelle Zhou. The machine-readable dictionary (BDC 1992) was provided by Behavior Design Corporation.